# POCS Based Super-Resolution Image Reconstruction Using an Adaptive Regularization Parameter

S.S.Panda[1], M.S.R.S Prasad[2] and Dr. G. Jena[3]

[1] Computer Science & Engg., Pondicherry University, Regency Institute of Technology
Yanam, Puducherry 533464, India
*shekhar_regency@yahoo.com*

[2] Computer Science & Engg., Pondicherry University, Regency Institute of Technology
Yanam, Puducherry 533464, India
*prasad_merla@yahoo.com*

[3] Computer Science & Engg., JNTU University, B V C Engineering College
Amalapuram, Andhra Pradesh 533210, India
*drgjena@rediffmail.com*

**Abstract**
Crucial information barely visible to the human eye is often embedded in a series of low-resolution images taken of the same scene. Super-resolution enables the extraction of this information by reconstructing a single image, at a high resolution than is present in any of the individual images. This is particularly useful in forensic imaging, where the extraction of minute details in an image can help to solve a crime. Super-resolution image restoration has been one of the most important research areas in recent years which goals to obtain a high resolution (HR) image from several low resolutions (LR) blurred, noisy, under sampled and displaced images. Relation of the HR image and LR images can be modeled by a linear system using a transformation matrix and additive noise. However, a unique solution may not be available because of the singularity of transformation matrix. To overcome this problem, POCS method has been used. However, their performance is not good because the effect of noise energy has been ignored. In this paper, we propose an adaptive regularization approach based on the fact that the regularization parameter should be a linear function of noise variance. The performance of the proposed approach has been tested on several images and the obtained results demonstrate the superiority of our approach compared with existing methods.

## 1. Introduction

Super-resolution image reconstruction is useful (sometimes essential) to enhance the image resolution in many practical cases such as medical imaging, satellite imaging, and video applications, where several low resolution images of the same content can be obtained. Signal processing techniques are often used to obtain HR images [1, 10]. In these techniques, the fusion of all kinds of information obtained from various LR images is used to enhance the resolution. The most successful methods are stochastic approaches in spatial domain such as ML (Maximum Likelihood), MAP (Maximum A-Posteriori), and POCS (Projection onto Convex Sets) [3]. In these methods, based on a linear model describing the relation of HR and LR images, a cost function is introduced and the HR image is estimated. POCS algorithm has been widely used because it is simple, can be applied to the occasion with any smooth movement, linear variable airspace vague and non-uniform additive noise and can easily join in the prior information. But POCS algorithm is strict to the accuracy of movement estimation. So in order to improve the stability of the algorithm, the relaxation operator will be used to replace ordinary projector operator, at the same time it is not conducive to the resumption of the edge and details of images [4]. However, the linear model used in these methods is an ill-posed problem in the sense that its transformation matrix may be singular and consequently a unique solution cannot be achieved [5,11]. In this paper, the problem of reconstructing an HR image is solved by choosing an adaptive regularization parameter to stabilize the inversion of ill-posed problem and to consider the effect of the noise. The obtained results show the performance of this new approach.

## 2. POCS Super-Resolution

K low resolution (LR) images can be seen as a high resolution (HR) after geometric distorting, fuzzy linear space, down sample and additional noise. Therefore, the imaging process can be written as [4]

$$G_k = D_k B_k M_k F_k + \xi_k \qquad (1)$$

$G_k$ is the observed image, size is M X M
$D_k$ is down sample matrix, size is M X L

$M_k$ is geometric distortion matrix, size is L X L
$B_k$ is fuzzy linear space matrix, size is L X L
F is the original high-resolution image, size is L X M
$\xi_k$ is Gaussian white noise, size is M ×M

Different observed images are obtained from different geometric distortion, spatial ambiguity, down sampling and additional noise to ideal images. The model above can be expressed as follows:

$$\begin{pmatrix} G_1 \\ . \\ . \\ . \\ G_N \end{pmatrix} = \begin{pmatrix} D_1 B_1 M_1 \\ . \\ . \\ . \\ D_1 B_1 M \end{pmatrix} * X + \begin{pmatrix} \xi_1 \\ . \\ . \\ . \\ \xi_N \end{pmatrix} = \begin{pmatrix} H_1 \\ . \\ . \\ . \\ H_N \end{pmatrix} * X + \xi_k \quad (2)$$

## 3. POCS Algorithm

The POCS approach to the SR reconstruction problem has been proposed in [7]. Let the motion information be provided. Then, a data consistency constraint set based on the acquisition model (2) can be defined for each pixel with each low-resolution (LR) image. The convex sets in any LR image are given by:

$$C_{n_1,n_2;i,k} = \left\{ x_i(m_1,m_2) : \left| r_k^{x_j}(n_1,n_2) \right| \leq \phi_0 \right\} \quad (3)$$

$0 \leq n_1, n_2 \leq N-1, k = 1,2,\ldots,L$

Where the value at each pixel is constrained such that it's associated residual

$$r_k^{x_i}(n_1,n_2) = g_k(n_1,n_2) - \sum_{m_1=0}^{M-1}\sum_{m_2=0}^{M-1} x_i(m_1,m_2) h_i(m_1,m_2;n_1,n_2) \quad (4)$$

is bounded in magnitude by $\varphi_0$ for the set. Since $\varphi_0$ is determined from the statistics of noise, the ideal image solution is a member of the set satisfying a certain statistical confidence. The projection of an arbitrary $x_i(m_1,m_2)$ onto $c_{n_1,n_2;i,k}$ is defined by

$$P_{n_1,n_2;j,k}[x_i(m_1,m_2)] = \begin{cases} x_i(m_1,m_2) + \frac{(r_k^{x_i}(n_1,n_2) - \phi_0) h_i(m_1,m_2;n_1,n_2)}{\sum_{m_1=0}^{M-1}\sum_{m_2=0}^{M-1} h_i^2(m_1,m_2;n_1,n_2)}, r_k^{x_i}(n_1,n_2) > \phi_0 \\ 0, -\phi_0 < r_k^{x_i}(n_1,n_2) < \phi_0 \\ x_i(m_1,m_2) + \frac{(r_k^{x_i}(n_1,n_2) + \phi_0) h_i(m_1,m_2;n_1,n_2)}{\sum_{m_1=0}^{M-1}\sum_{m_2=0}^{M-1} h_i^2(m_1,m_2;n_1,n_2)}, r_k^{x_i}(n_1,n_2) > -\phi_0 \end{cases} \quad (5)$$

Additional constraints such as bounded energy, positivity, and limited support may be utilized to improve the results. A generally utilized altitude constraint set is

$$P_A[x_i(m_1,m_2)] = \begin{cases} 0, x_i(m_1,m_2) < 0 \\ x_i(m_1,m_2), 0 \leq x_i(m_1,m_2) \leq 255 \\ 255, x_i(m_1,m_2) > 255 \end{cases}$$

## 4. Regularization Approach

One of the difficulties in solving Eq. (1) is inverting the forward model without amplifying the effect of noise. In addition, the minimizing process should make the solution insensitive to noise. Because the transformation matrix ( H ) is ill-conditioned, the obtained result may not be the best one. Thus some forms of regularization must be included in the cost function to solve the problem and to minimize the effect of the noise. The regularization is indeed some sort of constraints imposed on the space of possible solutions and is often independent of measured data [8].

$$J(X) = \|Y - HX\|^2 + \lambda p(X)$$

The function $p(X)$ poses a constraint on the unknown $X$ to direct it to a stable solution, and the coefficient $\lambda$ represents the strength of this constraint.

One of the most common forms for p(X) is Tikhonov regularization; $p(X) = \|TX\|^2$, in which $T$ is a matrix determined according to some aspects of the desired image such as energy or smoothness.

### 4.1 Proposed Algorithm

In this paper, we take T = I (identity matrix) which is minimal energy regularization and leads to a stable and unique solution [8].

$$J(X) = \|Y - HX\|^2 + \lambda \|X\|^2$$

Another important issue is the proper choice of $\lambda$. $\lambda$ is regularization parameter that controls a trade-off between the fidelity to data (expressed by $\|Y-HX\|^2$) and the energy of the solution (expressed by $\|X\|^2$). Larger values of $\lambda$ are useful when small number of LR images are available or the fidelity of data is low, which is caused by registration error and noise. On the other hand, smaller values of $\lambda$ are helpful whenever we have large number of LR images and noise is small [1]. As a result, larger noise variances lead to larger values of $\lambda$ and vice versa.

Based on above observation, we can conclude that there is a linear relation between $\lambda$ and noise variance as the form of $\lambda = m\sigma^2 + c$, where $\sigma^2$ is the noise variance, $m$ represents the weight of noise energy and $c$ is an offset. This is an adaptive regularization model that more efficiently minimizes Eq. (11) with respect to the effect of the noise.

It is important to note that although like regularization approaches an extra term is included in POCS estimator, because we do not have enough data to accurately estimate correlation matrix of *X, C* cannot be a precise estimation. It is the reason that our proposed regularization approach with adaptive regularization parameter can lead to better results.

## 5. Simulation Result

In our simulations LR images have been obtained by random displacement with uniform distribution over -10 and 10 pixels, blurring using linear motion by 5 pixels with an angle of 5 degrees, and the decimation with L1 =L2 = 2. Moreover, LR images have been corrupted by noise (AWGN) with SNR = 20 dB . Fig. 1 shows the original test image of Lena and its LR images.

To quantitatively evaluate the performance of the algorithms, we computed the PSNR using the original and reconstructed images. PSNR in dB is defined as [5]

$$PSNR = 10\log\left[\frac{255^2}{\frac{1}{N}\|\hat{X}-X\|^2}\right]$$

Where N is the total number of pixels, X is the original image, and $\hat{X}$ is the reconstructed image [9].

Firstly, in order to select the best linear model, many values of weight (m) and offset (c) have been tested. Table I shows the results on three well-known test images. The best results were achieved using m =$10^{-10}$ and c = 0.

Secondly, the performance of POCS and the proposed algorithm has been evaluated and compared using different test images. Table II shows PSNR values for different images using POCS, and proposed algorithm with m = $10^{-10}$ and c= 0.

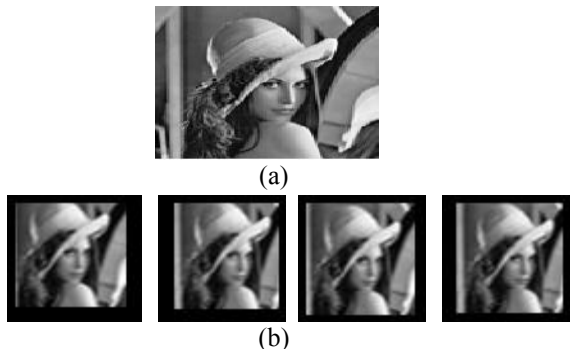

(a)

(b)

Figure 1 (a) original image (twice the real size), (b) four frames of LR images used to estimate the high resolution image.

Table I. Comparison Of Psnr Values For Different Images And Different Weights With c=0 (Psnr In Db)

| Tested images | Different values of m | | | | |
|---|---|---|---|---|---|
| | $10^{-1}$ | $10^{-3}$ | $10^{-5}$ | $10^{-10}$ | $10^{-12}$ |
| Lena | 8.65 | 28.22 | 67.39 | 72.54 | 71.54 |
| Peppers | 6.58 | 25.32 | 61.76 | 65.26 | 64.26 |
| Barbara | 8.10 | 29.74 | 66.37 | 69.20 | 68.20 |

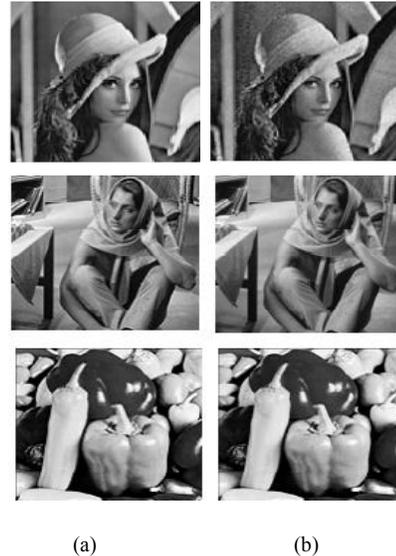

(a)       (b)

Figure 2. Column (a) shows the original images and column, Column (b) shows the reconstructed images using proposed algorithm for Lena, Barbara and Peppers .

Table II. Quantitative Performance Comparison Of POCS Algorithms Using PSNR

| Tested Image | Size | LR image | POCS | Proposed Algorithm |
|---|---|---|---|---|
| lena | 128x128 | 4 | 54.32 | 69.80 |
| Barbara | 128X128 | 4 | 55.92 | 69.95 |
| Peppers | 128X128 | 4 | 47.58 | 67.90 |
| Test 1 | 30X30 | 15 | 63.81 | 73.04 |
| Test 2 | 40X40 | 15 | 74.55 | 80.94 |
| Test 3 | 50X50 | 15 | 48.33 | 62.70 |
| Test 4 | 100X100 | 15 | 69.73 | 71.52 |

It can be seen that the proposed algorithm significantly improves the quality of the reconstructed high resolution images for all tested images compared with POCS algorithm.

## 4. Conclusions

This paper deals with achieving a high resolution image from several low resolution images of the same content. Specifically, we propose an adaptive regularization approach based on our observation that the noise energy should affect the regularization process. We have tested our approach on a variety of different images with different resolutions, and provided superior performance compared with other existing stochastic methods. The quantitative evaluation of the algorithms is based on PSNR which allows a good performance assessment.

**S. S. Panda** got his M.Tech. degree in the year 2010 and doing his research on Digital image processing. He is working as Sr. Assistant Professor in the department of Computer Science and Engineering, Regency Institute of Technology, Yanam.(Puducherry), India. He is a life member of ISTE.

**MSRS Prasad** got his M.Tech. degree in the year 2010 and doing his research on Digital image processing. He is working as Sr. Assistant Professor in the department of Computer Science and Engineering, Regency Institute of Technology, Yanam. (Puducherry), India. He is a life member of ISTE.

**Dr. G. Jena** got his Ph.D. degree from F M university and working as Professor and HOD CSE, BVC Engineering College, Amalapuram, India. His area of interest is Image Processing, Signal Processing and Adaptive Channel Equalizers using Time Frequency Domain Transform and Neural Techniques and is a member of the IEEE,FIE,LMISTE, MCSI, India .